\documentclass[10pt,twocolumn,letterpaper]{article}

\usepackage{cvpr}              % To produce the CAMERA-READY version
\usepackage[accsupp]{axessibility}
\usepackage{graphicx}
\usepackage{amsmath}
\usepackage{amssymb}
\usepackage{booktabs}
\usepackage{multirow}
\usepackage{bigstrut}
\usepackage{bm}
\usepackage{algorithm}
\usepackage[dvipsnames]{xcolor}
\usepackage{color}
\usepackage{algorithmicx}
\usepackage{algpseudocode}
\usepackage{colortbl}
\usepackage[american]{babel}
\usepackage{microtype}
\usepackage{threeparttable}

\usepackage[pagebackref,breaklinks,colorlinks]{hyperref}

\usepackage[capitalize]{cleveref}
\crefname{section}{Sec.}{Secs.}
\Crefname{section}{Section}{Sections}
\Crefname{table}{Table}{Tables}
\crefname{table}{Tab.}{Tabs.}

\def\cvprPaperID{8646} % *** Enter the CVPR Paper ID here
\def\confName{CVPR}
\def\confYear{2023}

\begin{document}

\title{Equiangular Basis Vectors}

\author{
Yang~Shen
\qquad  Xuhao Sun
\qquad  Xiu-Shen Wei\thanks{Corresponding author. This work was supported by National Key R\&D Program of China (2021YFA1001100), National Natural Science Foundation of China under Grant (62272231), Natural Science Foundation of Jiangsu Province of China under Grant (BK20210340), the Fundamental Research Funds for the Central Universities (No. NJ2022028), and CAAI-Huawei MindSpore Open Fund.} \\
{School of Computer Science and Engineering, Nanjing University of Science and Technology, China}\\
\small{\texttt{\{shenyang\_98,sunxh,weixs\}@njust.edu.cn}}
}

\maketitle

%%%%%%%%% ABSTRACT
\begin{abstract}
We propose Equiangular Basis Vectors~(EBVs) for classification tasks. In deep neural networks, models usually end with a $k$-way fully connected layer with \texttt{softmax} to handle different classification tasks. The learning objective of these methods can be summarized as mapping the learned feature representations to the samples' label space. While in metric learning approaches, the main objective is to learn a transformation function that maps training data points from the original space to a new space where similar points are closer while dissimilar points become farther apart. Different from previous methods, our EBVs generate normalized vector embeddings as ``predefined classifiers" which are required to not only be with the equal status between each other, but also be as orthogonal as possible. By minimizing the spherical distance of the embedding of an input between its categorical EBV in training, the predictions can be obtained by identifying the categorical EBV with the smallest distance during inference. Various experiments on the ImageNet-1K dataset and other downstream tasks demonstrate that our method outperforms the general fully connected classifier while it does not introduce huge additional computation compared with classical metric learning methods. Our EBVs won the first place in the 2022 DIGIX Global AI Challenge, and our code is open-source and available at \url{https://github.com/NJUST-VIPGroup/Equiangular-Basis-Vectors}.
   
\end{abstract}

%%%%%%%%% BODY TEXT
\section{Introduction} \label{sec:intro}

\begin{figure}[t]
\centering
{\includegraphics[width=0.5\textwidth]{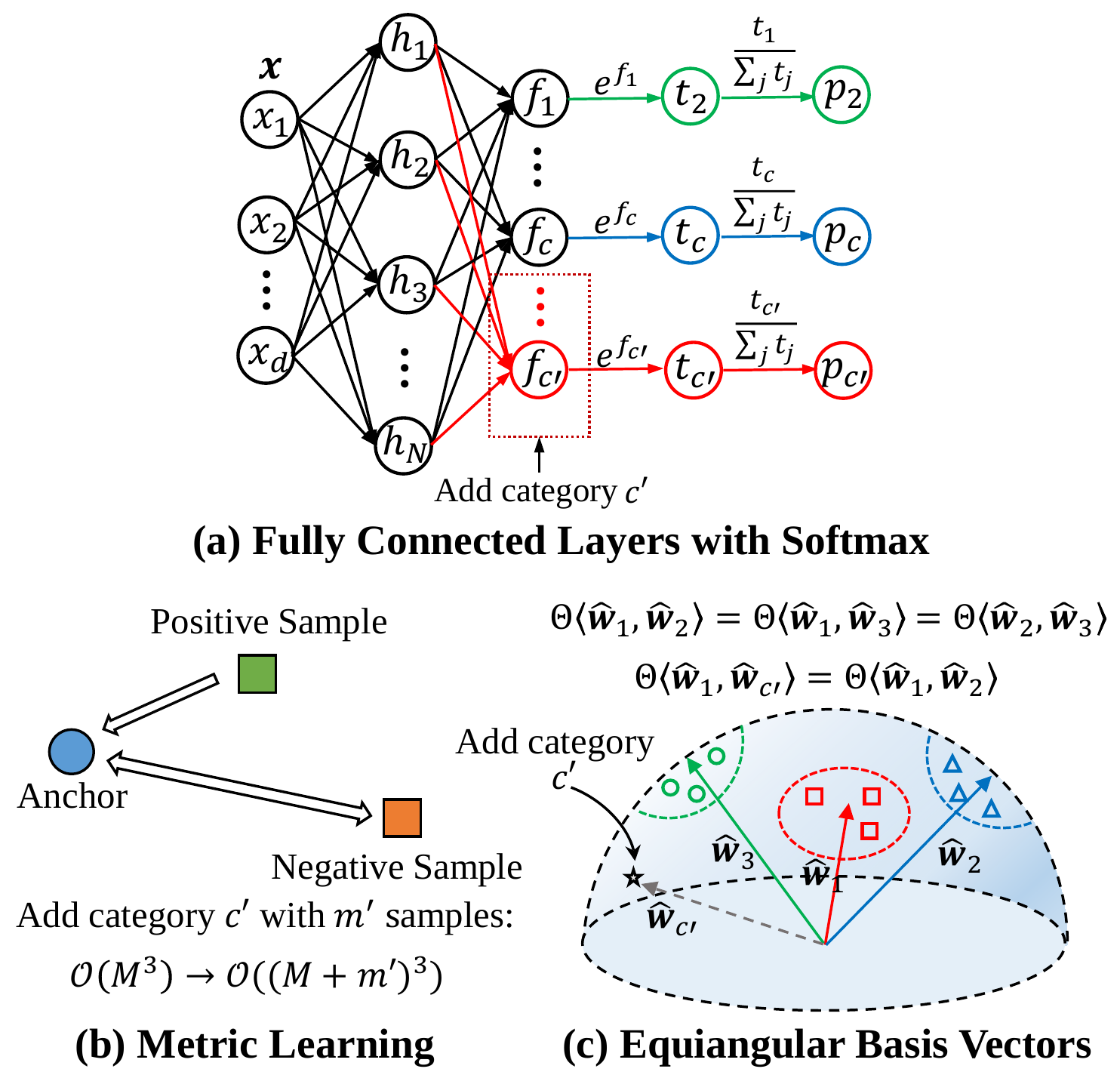}}
\vspace{-1.0em}
\caption{Comparisons between typical classification paradigms and our proposed Equiangular Basis Vectors~(EBVs). \textbf{(a)} A general classifier ends with $k$-way fully connected layers with \texttt{softmax}. When adding more categories, the trainable parameters of the classifier grow linearly. \textbf{(b)} Taking triplet embedding~\cite{wang2014learning} as an example of classical metric learning methods, the complexity is $\mathcal{O}(M^3)$ when given $M$ images and it will grows to $\mathcal{O}((M+m')^3)$ when adding a new category with $m'$ images. \textbf{(c)} Our proposed EBVs. EBVs predefine fixed normalized vector embeddings for different categories and these embeddings will not be changed during the training stage. The trainable parameters of the network will not be changed with the growth of the number of categories while the complexity only grows from $\mathcal{O}(M)$ to $\mathcal{O}(M+m')$.}
\label{fig: main}
\vspace{-1.0em}
\end{figure}

The pattern classification field developed around the end of the twentieth century aims to deal with the specific problem of assigning input signals to two or more classes~\cite{tulyakov2008review}. In recent years, deep learning models have brought breakthroughs in processing image, video, audio, text, and other data~\cite{krizhevsky2017imagenet,he2016deep,vaswani2017attention,dosovitskiy2020image}. Aided by the rapid gains in hardware, deep learning methods today can easily overfit one million images~\cite{deng2009imagenet} and easily overcomes the obstacle to the quality of handcrafted features in previous pattern classification tasks. Many approaches based on deep learning spring up like mushrooms and had been used to solve classification problems in various scenarios and settings such as remote sensing~\cite{maxwell2018implementation}, few-shot~\cite{snell2017prototypical}, long-tailed~\cite{zhou2020bbn}, etc. 

%such a classifier has poor scalability and
Figure~\ref{fig: main} illustrates two typical classification paradigms. Nowadays, a large amount of deep learning methods~\cite{maxwell2018implementation,zhou2020bbn} adopt a trainable fully connected layer with \texttt{softmax} as the classifier. However, since the number of categories is fixed, the trainable parameters of the classifier rise as the number of categories becomes larger. For example, the memory consumption of a fully connected layer $\bm{W}\in \mathbb{R}^{d\times N}$ linearly scales up with the growth of the category number $N$ and so is the cost to compute the matrix multiplication between the fully connected layer and the $d$-dimensional features. While some other methods based on classical metric learning~\cite{kaya2019deep,wen2016discriminative,lim2013robust,weinberger2009distance,wang2014learning} have to consider all the training samples and design positive/negative pairs then optimize a class center for each category, which requires a significant amount of extra computation for large-scale datasets, especially for those pre-training tasks.

In this paper, we propose Equiangular Basis Vectors (EBVs) to replace the fully connected layer associated with \texttt{softmax} within classification tasks in deep neural networks. EBVs predefine fixed normalized vector embeddings with equal status~(equal angles) which will not be changed during the training stage. Specifically, EBVs pre-set a $d$-dimensional unit hypersphere, and for each category in the classification task, EBVs assign the category a $d$-dimensional normalized embedding on the surface of the hypersphere and we term these embedding as \emph{basis vectors}. The spherical distance of each basis vector pair satisfies an artificially made rule to make the relationship between any two vectors as close to orthogonal as possible. In order to keep the trainable parameters of the deep neural networks constant with the growth of the category number $N$, we then propose the definition of EBVs based on Tammes Problem~\cite{tammes1930origin} and Equiangular Lines~\cite{van1966equilateral} in Section~\ref{sec: def}.

The learning objective of each category in our proposed EBVs is also different from previous classification methods. Compared with deep models that end with a fully connected layer to handle the classification tasks~\cite{krizhevsky2017imagenet,he2016deep}, the meaning of the parameter weights within the fully connected layer in EBVs is not the relevance of a feature representation to a particular category but a fixed matrix which embed feature representations to a new space. Also, compared with regression methods~\cite{salman2012regression,loh2011classification}, EBVs do not need to learn the unified representations for different categories and optimize the distance between the representation of input images and category centers, which helps reduce the computational consumption for the extra unified representations learning. In contrast to classical metric learning approaches~\cite{kaya2019deep,goldberger2004neighbourhood,salakhutdinov2007learning,weinberger2009distance}, our EBVs do not need to measure the similarity among different training samples and constrain distance between each category, which will introduce a large amount of computational consumption for large-scale datasets. In our proposed method, the representations of different images learned by EBVs will be embedded into a normalized hypersphere and the learning objective is altered to minimize the spherical distance of the learned representations with different predefined basis vectors. In addition, the spherical distance between each predefined basis vector is carefully constrained so that there is no need to spend extra cost in the optimization of these basis vectors. To quantitatively prove both the effectiveness and efficiency of our proposed EBVs, we evaluate EBVs on diverse computer vision tasks with large-scale datasets, including classification on ImageNet-1K, object detection on COCO, as well as semantic segmentation on ADE20K.

%------------------------------------------------------------------------
\section{Related work} \label{sec:formatting}
\subsection{Deep networks for image classification}
Image classification is the task of categorizing images into one of several predefined classes, which has been a fundamental problem in computer vision for a long time~\cite{chapelle1999support, lu2007survey}. It also forms the basis for many other computer vision tasks, \eg, object detection~\cite{malisiewicz2012exemplar}, localization~\cite{lampert2008beyond} and segmentation~\cite{mccallum2000maximum}. To solve the image classification problem, a dual-stage approach was used before the rise of deep learning. Specifically, handcrafted features were first extracted from images using feature descriptors. Then, a trainable classifier is adopted to perform the classification task with these input features~\cite{rawat2017deep}. The major hindrance of this approach was that the accuracy of the classification task was profoundly dependent on the design of the feature extraction stage, and this usually proved to be a formidable task~\cite{lecun1998gradient}.

In recent years, deep learning models that exploit multiple layers of nonlinear information processing, for feature extraction and transformation as well as for pattern analysis and classification, have been shown to overcome these challenges~\cite {rawat2017deep}. With the holding of ImageNet Large Scale Visual Recognition Challenge (ILSVRC)~\cite{russakovsky2015imagenet}, a growing number of deep networks demonstrate superior classification performance~\cite{krizhevsky2017imagenet,he2016deep,dosovitskiy2020image,radosavovic2020designing,liu2021swin,guo2022visual,liu2022convnet}. Among them, Deep Convolutional Neural Networks~(DCNNs) and Vision Transformers~(ViTs)~\cite{dosovitskiy2020image} have become the leading architectures for most image classification task in recent years. In addition, selected representative examples of other improvement attempts related to the following different aspects: network architecture, nonlinear activation functions, supervision components, regularization mechanisms and optimization techniques. Our proposed Equiangular Basis Vectors~(EBVs) are different from these aspects as we do not change the overall architecture and training techniques but the optimization objectives for classification since we preliminary design fixed normalized vector embeddings for each predefined category. 

\subsection{Learning objectives}
Besides traditional classifiers, machine learning or deep learning methods such as clustering~\cite{reimers2019classification,baker1998distributional}, regression~\cite{salman2012regression,loh2011classification}, metric learning~\cite{goldberger2004neighbourhood,salakhutdinov2007learning,weinberger2009distance}, sparse learning~\cite{rodriguez2008sparse,yuan2012visual} can be used to handle the classification tasks while the learning objectives of these methods vary considerably. In this section, we only discuss the learning objective of our EBVs with two prominent training paradigms of deep learning for the classification tasks, \ie, general $k$-way classification layers~\cite{krizhevsky2017imagenet} and classical deep metric learning~\cite{bellet2013survey,hanjia2019local}.

% As EBVs are concerned with supervised learning methods

Regarding deep learning for classification, the $k$-way classification layer (always associated with \texttt{softmax}) is the most popular used approach for training deep models. It employs a single linear layer or multiple non-linear layers to map the learned deep representations to the semantic categories in the label space. The corresponding learning objective is minimizing the losses (\eg, the cross-entropy loss~\cite{de2005tutorial}) between the mapped categorical signals (aka. predictions) with the ground truth categories. Compared with that, the learning objective of our EBVs is to make the vectorized embedding of input as close as possible to its categorical equiangular basis vector correspondingly. 

The learning objective of classical deep metric learning seems similar to ours. However, there are several crucial and fundamental differences. Firstly, although deep metric learning minimizes the distances between sample vectors belonging to the same category and meanwhile maximizes the distances between samples from different categories, all these sample vectors are constantly changing during the model training process. But, in our method, the equiangular basis vectors corresponding to the categories are predefined, \ie, they are fixed. More importantly, these equiangular basis vectors in the spherical space are forced to be with equal status and be as orthogonal as possible to each other, which could contribute to strong model discriminative ability and good classification accuracy. In practice, both with equal status and being orthogonal of vectors are strict conditions w.r.t. optimization. Thus, our EBVs predefine the satisfied categorical basis vectors and our learning objective focuses on only optimizing the learned feature representations to prevent the model from being unable to converge (when EBVs can also change under such strict conditions). Secondly, the learning objective of deep metric learning is upon massive training samples, while our learning objective is for the training sample and its fixed categorical vectorized embedding, \ie, EBV. It could bring the computational economy, especially for the large-scale training data scenario, cf. Section~\ref{sec:merits}. In addition, compared with several specific metric learning approaches, \eg, the center loss~\cite{wen2016discriminative}, the prototypical network~\cite{snell2017prototypical} and the nearest class mean approach~\cite{mensink2013distance}, our method is still quite distant from them. Specifically, the basic idea of these works is to construct the ``center'' of samples belonging to a category to represent its semantics and then leverage the centers to optimize sample distances for classification. Similarly, since these categorical centers are always changing during training, these approaches will also encounter the same problems as aforementioned.

\section{Methodology}
\subsection{Preliminaries} \label{sec: pre}
The proposed Equiangular Basis Vectors~(EBVs) are based on the study with regard to Equiangular Lines~\cite{Haa48,van1966equilateral,jiang2021equiangular} and the Tammes Problem~\cite{tammes1930origin}. 

A set of lines passing through the origin in $\mathbb{R}^d  (d\in \{2,3,4,5,\ldots\})$ is called \emph{equiangular} if they are pairwise separated by the same angle. The study of equiangular lines was initiated by Haantjes~\cite{Haa48} and it plays an important role in the coding theory~\cite{strohmer2003grassmannian} and quantum information theory~\cite{renes2004symmetric}. 

The problem of how to determine the maximum number $N(d)$ of equiangular lines in a given dimension $d$ was formally posed by Van Lint and Seidel~\cite{van1966equilateral} and has gained a major breakthrough by Jiang et al.~\cite{jiang2021equiangular} last year — Fix $0 < \alpha < 1$, let $N_\alpha(d)$ denote the maximum number of lines through the origin in $\mathbb{R}^d$ with pairwise common angle $\arccos \alpha$, let $k$ denote the minimum number (if it exists) of vertices in a graph whose adjacency matrix has spectral radius exactly $(1-\alpha)/(2\alpha)$. If $k < \infty$, then $N_\alpha(d) = \lfloor k(d-1)/(k-1) \rfloor$ for all sufficiently large $d$, and otherwise $N_\alpha(d) = d + o(d)$. In particular, $N_{1/(2k-1)}(d) = \lfloor k(d-1)/(k-1) \rfloor$ for every integer $k\ge 2$ and all sufficiently large $d$~\cite{jiang2021equiangular}.

In simple terms, with a fixed common angle, the maximum number $N(d)$ of equiangular lines is linearly correlated with the dimension $d$ as $d \to \infty$ while there is still no precise lower bound for $N(d)$ with smaller values of $d$~\cite{glazyrin2018upper}. Therefore, we further refer to the Tammes Problem. Let $a_n$ be the maximal number with the property that one can place on a unit hypersphere $S^{d} \in \mathbb{R}^d (d\in \{ 3,4,5,\ldots\})$ $n$ points so that the spherical distance of any two points is at least $a_n$. The problem of finding $a_n$ together with the corresponding arrangement of each point is known as the Tammes Problem~\cite{tammes1930origin} or the optimal spherical code~\cite{ericson2001codes}. It is easy to find that the number of these points will be close to infinity as $a_n$ tends to zero.

\subsection{Definition of Equiangular Basis Vectors} \label{sec: def}
In order to predefine fixed $d$-dimensional embeddings for as many categories as possible but still keep these embeddings at a distance from each other on a unit hypersphere $S^{d} \in \mathbb{R}^d$, we therefore propose Equiangular Basis Vectors~(EBVs). Specifically, we fix $0 \leq \alpha < 1$ and let $\mathcal{N}_\alpha(d)$ denote the range of values for the number of coordinate vectors in $\mathbb{R}^d$ with the pairwise common angle between $\arccos \alpha$ and $\arccos -\alpha$. The problem of our proposed EBVs is to calculate the coordinates of each vector in the vector set $\mathcal{W}$ when given fixed $\alpha$, $d$ and $N \in \mathcal{N}_\alpha(d)$ if possible, \ie, solve the set $\mathcal{W}$ which satisfies:
\begin{equation}\label{eq:def_ebv}
    \forall \bm{w}_i, \bm{w}_j \in \mathcal{W}, i\ne j, \quad 
    -\alpha \leq \frac{\bm{w}_i \cdot \bm{w}_j}{\left \| \bm{w}_i \right \| \left \| \bm{w}_j \right \|} \leq \alpha\,,  
\end{equation}
where $\bm{w}_i \in \mathbb{R}^{d}$, ${\rm card}(\mathcal{W})= N$ and $\left \| \cdot \right \| $ denotes the Euclidean norm. Let $\phi$ denote the spherical distance function, which can also be replaced by the cosine similarity function. EBVs produce a distribution over classes for a query point $\bm{v}\in \mathbb{R}^d$ based on \texttt{softmax} over cosine similarity to the $N$ fixed coordinate vectors in the embedding space:
\begin{equation}\label{eq:3}
    p(y=k|\bm{v})=\frac{{\rm exp}(-\phi(\bm{v},\bm{w}_k))}{\sum_{k'} {\rm exp}(-\phi(\bm{v}, \bm{w}_{k'}))}\,,
\end{equation}
where $y=k\in \{1,2,\ldots, N\}$ denotes the corresponding coordinate vector, which can also be seen as the corresponding label. While $k'$ represents the associated basis vectors in $\mathcal{W}$. Relations between $\alpha$, $d$ and $N$ are described in the supplementary materials. With such a set $\mathcal{W}$, the maximum number of categories it can handle is $N$. Additionally, training with any category number less than $N$, it is sufficient to randomly select any of the same numbers of coordinate vectors in $\mathcal{W}$, since these vectors are exactly equivalent.

\subsection{How to generate EBVs?}\label{sec:GEBV}
The basic idea of our Equiangular Basis Vectors~(EBVs) is to generate fixed normalized vector embeddings with equal status~(equal angles) as ``predefined classifiers" for all the relevant categories. The question then comes to how to calculate the predefined EBVs which satisfy Eq.~\eqref{eq:def_ebv}. Therefore, we will discuss how to generate the proposed EBVs when given fixed $\alpha$, $d$ and $N \in \mathcal{N}_\alpha(d)$ in this section.

Assuming each $\bm{w}_i \in \mathcal{W}\ (i=\{1,2,\ldots, N \})$ as a line, we can construct the \emph{Grassmannian Matrices} to solve the set $\mathcal{W}$~\cite{elad2010sparse}. Specifically, we assemble the vectors in $\mathcal{W}$ into a matrix $\bm{W}\in \mathbb{R}^{d\times N}$, where $\bm{W}=[\bm{w}_1;\bm{w}_2;\ldots;\bm{w}_N]$. Then, the mutual-coherence of $\bm{W}$ is defined by:
\begin{equation}\label{eq:4}
    \mu (\bm{W}) = \mathop{\rm max}\limits_{1\le i,j\le N, i\ne j} \frac{\left | \bm{w}_i^{\top}\cdot \bm{w}_j  \right |}{\left \| \bm{w}_i \right \| \left \| \bm{w}_j \right \|} = \mathop{\rm max}\limits_{1\le i,j\le N, i\ne j}{\rm cos}\ \theta_{ij}\,,
\end{equation}
which is transformed to the smallest possible mutual-coherence possible~\cite{elad2010sparse}. Therefore, the lower bound for $\alpha$ is $\sqrt{\frac{N-d}{d(N-1)}}$ and the upper bound for $\mathcal{N}_\alpha(d)$ is $1 + \frac{d-1}{1-\alpha^2d}\ (d\le N, 1 - \alpha^2d > 0)$. In addition, such matrix is possible only if $N < {\rm min}(d(d+1)/2, (N-d)(N-d+1)/2)$. Construction of such a matrix has a strong connection with the packing of vectors/subspaces in the $\mathbb{R}^d$-space. In the case of $N=d$, we can simply construct a unitary matrix, while in other cases, it will be very hard to construct such a general Grassmannian matrix~\cite{elad2010sparse}. 

In the definition of our proposed EBVs, whether the angle between any two basis vectors $\bm{w}_i, \bm{w}_j \in \mathcal{W}\ (i\ne j)$ is $\arccos \alpha$ or $\arccos -\alpha$ is equivalent. It is clear that we can not construct the Grassmannian matrix in a situation such as $N=30,000$ and $d=200$. However, it is still possible for us to construct $\mathcal{W}$ which satisfies Eq.~\eqref{eq:def_ebv}. Therefore, as an alternative, we adopt Stochastic Gradient Descent~\cite{robbins1951stochastic} to search the set $\mathcal{W}$ that satisfies the definition of EBVs when given fixed $\alpha$, $d$ and $N$. Specifically, we random initialize a matrix $\bm{W}\in \mathbb{R}^{d\times N}$ with normalized rows such that the angle between any two vectors $\hat{\bm{w}}_i$, $\hat{\bm{w}}_j\in \mathbb{R}^{d}$, $i,j\in \{1,2,\ldots,N\},i\neq j, \hat{\bm{w}}_i =\frac{\bm{w}_i}{\left \| \bm{w}_i \right \|}$ can be represented as $\arccos (\hat{\bm{w}}_i \cdot \hat{\bm{w}}_j)$. Then, we cut out the gradient of those vector pairs which satisfies $-\alpha \leq \hat{\bm{w}}_i \cdot \hat{\bm{w}}_j \leq \alpha$ and optimize the remaining vector pairs. The optimization function of the generation of EBVs can be formulated by:
\begin{equation}\label{eq:5}
    \mathop{\arg \min}_{\bm{W}} \sum\nolimits_{i=1}^{N-1} \sum\nolimits_{j > i}^{N} {\rm max} (\hat{\bm{w}}_i \cdot \hat{\bm{w}}_j - \alpha , 0)\,. 
\end{equation}
Algorithm~\ref{alg:a1} provides the code of a simple generation method of the proposed EBVs in a PyTorch-like style. It is also worth mentioning that the EBVs matrix $\bm{W}$ will not be changed in the following training stage within all the tasks.

\begin{algorithm}[t]
\footnotesize
	\caption{Generation of EBVs in a PyTorch-like style}
	\label{alg:a1}
	
	\textcolor{teal}{\# $d$: Dim for each coordinate vectors\;} 
	
	\textcolor{teal}{\# $N$: The number of coordinate vectors\;}
	
	\textcolor{teal}{\# $\alpha$: Threshold of $\hat{\bm{w}}_i \cdot \hat{\bm{w}}_j$, $i\ne j$\;} 
	
	\textcolor{teal}{\# $\bm{W}$: The EBVs matrix $\bm{W} \in \mathbb{R}^{d\times N}$\;}
	
	\textcolor{teal}{\# slice: In the case of $N \gg d$, optimize $\bm{W}$ by slicing\;}
	
	\textcolor{teal}{\# $\lambda$: Learning rate.}\\
	%\KwOut{Vector set $\mathcal{W}'$ which satisfies Eq.~\eqref{eq:def_ebv}.}  
	
	Initialize $\bm{W}$ randomly;
	
	while True:
  
    \qquad \textcolor{teal}{\#Normalize each row in $\bm{W}$}
    
    \qquad $\bm{W}$ = Normalize($\bm{W}$)
  
    \qquad for i in $\left \lceil N / slice \right \rceil $:
    
    \qquad \qquad start = i * slice
    
    \qquad \qquad end = ${\rm min}$($N$, (i $+$ 1) * slice)
    
    \qquad \qquad E = F.onehot(arange(start, end), $N$)
    
    \qquad \qquad C = (W[start:end]@W.T).abs()-E 
    
    \qquad \qquad loss = ${\rm ReLU}$(C $-$ $\alpha$).sum() \textcolor{teal}{\# Cutout the} 
    
    \qquad \qquad \textcolor{teal}{gradient of vector pairs which satisfies} 
    
    \qquad \qquad \textcolor{teal}{$-\alpha \leq \hat{\bm{w}}_i \cdot \hat{\bm{w}}_j \leq \alpha$}
    
    \qquad \qquad loss.backward()
    
    \qquad if ${\rm max}(\alpha$, C.${\rm max}())$ $<$ $\alpha + o(\alpha)$:
    
    \qquad \qquad Save($\bm{W}$)
    
    \qquad \qquad break
  
    \qquad $\bm{W}$ = $\bm{W}$ $-$ $\lambda$ * $\bm{W}$.grad \textcolor{teal}{\# Update $\bm{W}$}
    
\end{algorithm}

\subsection{How to achieve the learning objective of EBVs?}
Equiangular Basis Vectors~(EBVs) provide fixed learning targets for each independent optimization objective, \ie, semantic categories. In the following, we introduce how to achieve the learning objective of EBVs for all the training samples. Generally, a deep network is performed to extract the high-dimensional features, and a fully connected classification layer is then deployed to map the features to semantic categories. While in our proposed EBVs, each category will be bound to a unique normalized $d$-dimensional basis vector in $\mathcal{W}$. Thus, for a training sample $x$, we directly use a unified deep model to generate a $d$-dimensional embedding $\bm{v}$, as well as optimizing the cosine distance between $\bm{v}$ and the relevant basis vector. Below we analyze the underlying distance/loss function used in the training stage.

For our EBVs, many existing distance functions including squared Euclidean distance, Mahalanobis distance, or cosine distance are permissible. However, a particular class of distance functions, \eg, \emph{regular Bregman divergences}~\cite{banerjee2005clustering}, seems hard to explain and optimize in the proposed EBVs settings. In addition, intuitively, the straightforward way is to optimize the spherical distance between any two vectors on the surface of the hyper-sphere. Therefore, we adopt the cosine distance as the distance metric, which is widely used for measuring whether two inputs are similar or dissimilar and have been widely used in the Tammes problem~\cite{tammes1930origin}.

\paragraph{Implementations} 
Suppose we have $M$ sample-label pairs $\{(x_1,y_1), (x_2,y_2), \ldots, (x_M,y_M)\}$ in $N$ classes and their $d$-dimensional features $\bm{v}_1, \bm{v}_2, \ldots, \bm{v}_M$ with $\bm{v}_i = f_{\bm{\theta}}(x_i)$, where $f_{\bm{\theta}}(\cdot)$ represents a feature extractor. A straightforward way to achieve the learning objective of EBVs is to directly optimize the cosine similarity between each $\bm{v}_i$ and the corresponding basis vector $\hat{\bm{w}}_{y_i}\in \mathcal{W}$. 

However, the basis vector itself is not correlated with the training data since it is predefined. Fitting the training samples directly to the corresponding basis vectors might disrupt the representation learning. Therefore, we consider the conventional parametric \texttt{softmax} formulation, of which for image $x_i$ with embedding $\bm{v}_i$, the probability of it being recognized as the $y_i$ category can be formulated as :
\begin{equation}\label{eq:sfc}
    P(y = y_i|\bm{v}_i)=\frac{{\rm exp}(\bm{m}_i^{\top} \bm{v}_i)}{\sum_{j=1}^{N} {\rm exp}(\bm{m}_j^{\top} \bm{v}_i)}\,,
\end{equation}
where $\bm{m}_j$ is a weight vector for the $j$-th class. Thus, according to Eq.~\eqref{eq:3} and Eq.~\eqref{eq:sfc}, the probability of the embedding $\bm{v}_i$ being recognized as category $y_i$ in our proposed EBVs can be formulated as:
\begin{equation}
    P_{\rm EBVs}(y = y_i|\bm{v}_i)=\frac{{\rm exp}(\hat{\bm{w}}_i \hat{\bm{v}}_i /\tau)}{\sum_{j=1}^{N} {\rm exp}(\hat{\bm{w}}_j \hat{\bm{v}}_i /\tau)}\,,
\end{equation}
where $\hat{\bm{v}}_i$ denotes the $\ell_2$-normalized of the embedding $\bm{v}_i$ and $\tau$ is a hyper-parameter of temperature~\cite{hinton2015distilling,wu2018unsupervised,he2020momentum} which usually used in unsupervised learning. The learning objective is then to maximize the joint probability ${\textstyle \prod_{i=1}^{M}}P_{\bm{\theta}} (y_i|f_{\bm{\theta}}(x_i))$, or equivalently to minimize the negative log-likelihood over the training set which can be formulated as:
\begin{equation}
    J(\bm{\theta}) = -\sum\nolimits_{i=1}^{M}{\rm log} P_{\rm EBVs}(y=y_i|f_{\bm{\theta}}(x_i))\,. 
\end{equation}
With this optimization approach, we relax the learning objective to make the angle between the embedding $\bm{v}$ of a training sample and its corresponding basis vector smaller than the angle between $\bm{v}$ and the other basis vectors.

\subsection{Merits of our EBVs}\label{sec:merits}
We briefly summarize the merits of our proposed EBVs in this section. First of all, the embedding dimension of EBVs can be manually altered and the trainable parameters of the classifier will not grow linearly when the number of categories increases. Specifically, if the embedding dimension of each image is $d'$ and the number of categories is $N$, the trainable parameters of a general classifier are $d'\times N$. However, with our proposed EBVs, as the dimension of the embedding for each category is fixed as $d$, then the trainable parameters of the classifier are $d'\times d$. What's more, $d$ can be set at least close to $\sqrt{2N}$ when $N$ is very large according to Section~\ref{sec:GEBV}. Experimental results can be found in Table~\ref{tab:EBV_dim} and the supplementary materials. 

Secondly, as the proposed EBVs are generated before the training step and the fixed $d$-dimensional embedding of each category will not be changed throughout the optimization process, EBVs will not introduce a large amount of computation during the training stage like the metric learning. In particular, metric learning methods such as pairwise/triplet embedding~\cite{wang2014learning} are time-consuming, where the complexity can be $\mathcal{O}(K^2)$ and $\mathcal{O}(K^3)$ when given $K$ images from $N$ categories. However, the complexity of our proposed EBVs equals $\mathcal{O}(K)$. In addition, EBVs are not sensitive to the optimizers and previous training tricks while they can still achieve state-of-the-art performance as the results shown in Table~\ref{tab:EBV_base} and the supplementary materials. Also, further results on the downstream tasks can be found in Section~\ref{sec: exp} and the supplementary materials.

%%%%%%%%%%%%-----------ImageNet-------------%%%%%%%%%%%
\begin{table*}[thbp]
  \centering
  \setlength\tabcolsep{1.8mm}
  \small
  \caption{Comparisons on the ImageNet-1K validation set. ``FC" denotes models ending with a $1000$-way fully connected layer with \texttt{softmax}. The test size for each image is set as ``$224^2$" if there is one result while it is set as ``$224^2$/$256^2$" if there exist two results.
  }
  \begin{threeparttable}
  \small
    \begin{tabular}{c|ccccccccc|c}
    \toprule
    Backbone & Method & Optimizer & LR    & Epoch & Setting & Params. & GFLOPs & Test size & \#Forward pass & Top-1 Acc ($\%$) \\
    \hline
    ResNet-50 & FC    & SGD   & 0.1   & 90    & A0     & 25.6M  & 4.1   & 224$^2$   & 600k  & 77.15 \bigstrut[t]\\
    \hline
    \multirow{2}[2]{*}{ResNet-50} & FC    & \multirow{2}[2]{*}{SGD} & \multirow{2}[2]{*}{0.5} & \multirow{2}[2]{*}{5/100} & \multirow{2}[2]{*}{A1} & \multirow{2}[2]{*}{25.6M} & \multirow{2}[2]{*}{4.1} & \multirow{2}[2]{*}{224$^2$/256$^2$} & \multirow{2}[2]{*}{131k} & 76.88/77.92 \bigstrut[t]\\
          & EBVs   &       &       &       &       &       &       &       &       & \textbf{77.55}/\textbf{78.73} \\
    \hline
    \multirow{2}[2]{*}{ResNet-50} & FC & \multirow{2}[2]{*}{SGD} & \multirow{2}[2]{*}{0.5} & \multirow{2}[2]{*}{5/100} & \multirow{2}[2]{*}{A2} & \multirow{2}[2]{*}{25.6M} & \multirow{2}[2]{*}{4.1} & \multirow{2}[2]{*}{224$^2$/256$^2$} & \multirow{2}[2]{*}{131k} & 77.71/78.59 \bigstrut[t]\\
          & EBVs   &       &       &       &       &       &       &       &       & \textbf{78.14}/\textbf{78.99} \\
    \hline
    \multirow{2}[2]{*}{ResNet-50*} & FC    & \multirow{2}[2]{*}{SGD} & \multirow{2}[2]{*}{0.5} & \multirow{2}[2]{*}{5/600} & \multirow{2}[2]{*}{A2} & \multirow{2}[2]{*}{25.6M} & \multirow{2}[2]{*}{4.1} & \multirow{2}[2]{*}{224$^2$/256$^2$} & \multirow{2}[2]{*}{755k} & 79.51/ -- \bigstrut[t]\\
          & EBVs   &       &       &       &       &       &       &       &       & \textbf{79.73}/\textbf{80.45} \\
    \hline
    \multirow{3}[2]{*}{ResNet-50} & FC    & \multirow{3}[2]{*}{AdamW} & 0.001 & \multirow{3}[2]{*}{5/100} & \multirow{3}[2]{*}{A1} & \multirow{3}[2]{*}{25.6M} & \multirow{3}[2]{*}{4.1} & \multirow{3}[2]{*}{224$^2$/256$^2$} & \multirow{3}[2]{*}{131k} & 72.57/73.79 \bigstrut[t]\\
          & FC    &       & 0.01 &       &       &       &       &       &       & 72.51/74.03 \\
          & EBVs   &       & 0.01 &       &       &       &       &       &       & \textbf{75.62}/\textbf{77.14} \\
    \hline
    \multirow{3}[2]{*}{ResNet-50} & FC    & \multirow{3}[2]{*}{AdamW} & 0.001 & \multirow{3}[2]{*}{5/100} & \multirow{3}[2]{*}{A2} & \multirow{3}[2]{*}{25.6M} & \multirow{3}[2]{*}{4.1} & \multirow{3}[2]{*}{224$^2$/256$^2$} & \multirow{3}[2]{*}{131k} & 75.42/76.48 \bigstrut[t]\\
          & FC    &       & 0.01 &       &       &       &       &       &       & NaN \\
          & EBVs   &       & 0.01 &       &       &       &       &       &       & \textbf{76.46}/\textbf{77.52} \\
    \hline
    \multirow{2}[2]{*}{Swin-T} & FC    & \multirow{2}[2]{*}{AdamW} & \multirow{2}[2]{*}{0.001} & \multirow{2}[2]{*}{5/100} & \multirow{2}[2]{*}{A1} & \multirow{2}[2]{*}{28.3M} & \multirow{2}[2]{*}{4.5} & \multirow{2}[2]{*}{224$^2$} & \multirow{2}[2]{*}{131k} & 75.64 \bigstrut[t]\\
          & EBVs   &       &       &       &       &       &       &       &       & \textbf{78.37} \\
    \hline
    \multirow{2}[2]{*}{Swin-T} & FC    & \multirow{2}[2]{*}{AdamW} & \multirow{2}[2]{*}{0.001} & \multirow{2}[2]{*}{5/100} & \multirow{2}[2]{*}{A2} & \multirow{2}[2]{*}{28.3M} & \multirow{2}[2]{*}{4.5} & \multirow{2}[2]{*}{224$^2$} & \multirow{2}[2]{*}{131k} & 79.12 \bigstrut[t]\\
          & EBVs   &       &       &       &       &       &       &       &       & \textbf{79.34} \\
    \bottomrule
    \end{tabular}%
    \begin{tablenotes}
        \item[*] represents the result of the model with a general fully connected layer with \texttt{softmax} is provided by TorchVision~\cite{va2021how}.
      \end{tablenotes}
  \end{threeparttable}
  \label{tab:EBV_base}%
\end{table*}%

\section{Experiments} \label{sec: exp}
In this section, quantitative and qualitative experiments are exhibited on models that end with a $k$-way fully connected layer with \texttt{softmax} and our proposed method to demonstrate the effectiveness of the proposed Equiangular Basis Vectors~(EBVs). All experiments are conducted with 8 RTX 3090 GPUs.

\subsection{Quantitative results on image classification}
\subsubsection{Dataset and settings}
We conduct the general image classification task on the ImageNet-1K~\cite{deng2009imagenet} dataset, which contains 1.28M training images and 50K validation images from 1,000 different object classes. Then, we report ImageNet-1K top-1 accuracy on the validation set under a single crop setting.

In order to demonstrate the effectiveness of the proposed EBVs, we follow the state-of-the-art training methods provided by TorchVision~\cite{va2021how} and timm~\cite{wightman2021resnet}. For fair comparisons, we offer the following diverse training settings for the general classification task. 

\paragraph{Setting A0} The offical training setting provided in ResNet~\cite{he2016deep}. A $224\times 224$ crop is randomly sampled from an image or its horizontal flip, with the per-pixel mean subtracted~\cite{krizhevsky2017imagenet}. It trains the backbone by using SGD optimizer with momentum as $0.9$, weight decay as $1\times 10^{-4}$ and batch size as $256$.  The training iteration is up to $60\times 10^4$. The standard color augmentation in~\cite{krizhevsky2017imagenet} is used. It adopts the standard \textbf{10-crop testing}~\cite{krizhevsky2017imagenet} in the validation stage. 

\paragraph{Setting A1} We employ an AdamW~\cite{loshchilov2017decoupled} or an SGD~\cite{robbins1951stochastic} optimizer for 100 epochs using a cosine decay learning rate scheduler and 5 epochs of linear warm-up. The batch size is set as $1024$ and the initial learning rate for the AdamW optimizer is $0.01$ or $0.001$ while it is set as $0.5$ for the SGD optimizer. A $224\times 224$ crop is randomly sampled from each image and the weight decay is set as $2\times 10^{-5}$. We perform TrivialAugment~\cite{muller2021trivialaugment}, which is extremely simple and can be considered ``parameter-free”. We also adopt random erasing~\cite{zhong2020random} and the probability is set as $0.1$.

\paragraph{Setting A2} On the basis of Setting A1, we add label-smoothing~\cite{szegedy2016rethinking} and the value is set as $0.1$. We also perform mixup~\cite{zhang2017mixup} and cutmix~\cite{yun2019cutmix}. The setting of hyper-parameters of the two techniques is the same with TorchVision~\cite{va2021how}.

We adopt ResNet-50~\cite{he2016deep} and Swin-T~\cite{liu2021swin} as the typical backbone for Convolutional Neural Networks~(CNNs) and Vision Transformers~\cite{dosovitskiy2020image}. The hyper-parameters $\tau$ for the proposed EBVs are set as $0.07$ for all the following experiments unless otherwise specified.

\subsubsection{Main results} 
Table~\ref{tab:EBV_base} presents comparisons between deep networks ending with FC, \ie, a 1000-way classification layer with \texttt{softmax} and our proposed EBVs. The fixed $\alpha$, $d$ and $N$ of EBVs are set as $0.004$, $1000$ and $1000$, respectively. When adopting ResNet-50 as the backbone, EBVs outperform FC among all the settings. In addition, when adopting AdamW as the optimizer, EBVs gain 3.11\%/1.04\% improvement over FC under Setting~A1 and Setting~A2 when using $224^2/256^2$ testing. EBVs still gain around 2.7\%/0.2\% improvement on Swin-T under Setting~A1 and Setting~A2. As previous training techniques summarised by both TorchVision~\cite{va2021how} and timm~\cite{wightman2021resnet} are under the setting of FC, we suspect it will have some impact on the performance of EBVs. We further conduct ablation studies in the supplementary materials.

%%%%%%%%%%%%-----------EBVs Dim-------------%%%%%%%%%%%
\begin{table}[t]
  \centering
  \renewcommand{\arraystretch}{1.0}
  \setlength\tabcolsep{0.2mm}
  \caption{Ablation studies on the dimension of our proposed EBVs. ``EBVs Dim." denotes the embedding dimension for each category.  The test size for each image is set as $224^2$ and $256^2$.
  %Min. Ang. denotes minimum angle among all basis vector pairs. Ave. Del. Ang. denotes the average delta angle among all basis vector pairs.
  }
    \begin{tabular}{c|ccc|c}
    \toprule
    EBVs Dim. & \multicolumn{1}{c}{Epoch} & Min. Ang. & Ave. Del. Ang. & \multicolumn{1}{c}{Top-1 Acc ($\%$)} \\
    \hline
    100   & 5/100 & 82.24$^\circ$ & 4.84$^\circ$  & 76.10/76.83 \bigstrut[t]\\
    200   & 5/100 & 85.41$^\circ$ & 3.37$^\circ$  & 77.04/77.83 \\
    300   & 5/100 & 86.67$^\circ$ & 2.64$^\circ$  & 77.25/78.08 \\
    400   & 5/100 & 87.48$^\circ$ & 2.18$^\circ$  & 77.20/78.38 \\
    500   & 5/100 & 87.99$^\circ$ & 1.82$^\circ$  & 77.39/78.34 \\
    1000  & 5/100 & 89.98$^\circ$ & 0.01$^\circ$  & \textbf{78.14}/\textbf{78.99} \\
    2000  & 5/100 & 89.98$^\circ$ & 0.01$^\circ$  & 77.87/78.58 \\
    \hline
    100   & 5/300 & 82.24$^\circ$ & 4.84$^\circ$  & 78.33/79.25 \bigstrut[t]\\
    1000  & 5/300 & 89.98$^\circ$ & 0.01$^\circ$  & \textbf{79.10}/\textbf{79.96} \\
    \bottomrule
    \end{tabular}%
  \label{tab:EBV_dim}%
\end{table}%

\subsubsection{Ablation studies} 
We conduct ablation studies on the dimension $d$ of each basis vector to demonstrate the merits of our proposed EBVs. Table~\ref{tab:EBV_dim}
reports ImageNet-1K top-1 accuracy on the validation set with different dimensions $d$. Taking $d=100$ as an example, the dimension of the embedding of an image is set as $100$. We adopt ResNet-50 as the backbone and the setting of all the experiments follows Setting A2. We use SGD~\cite{robbins1951stochastic} as the optimizer and the initial learning rate is set as $0.01$. The test size of each image is set as $224^2$ and $256^2$. The Min. Ang. which represents the minimum angle between each two basis vectors is defined as $\min_{1\le i,j \le N, i\ne j} \min (\pi - \arccos(\hat{\bm{w}}_i \cdot \hat{\bm{w}}_j), \arccos(\hat{\bm{w}}_i \cdot \hat{\bm{w}}_j))$, where $N$ represents the number of categories. While the Ave. Del. Ang. which represents the average angle between each of two basis vectors is defined as $\frac{N(N-1)}{2} \sum_{i=1}^{N-1} \sum_{j=i+1}^{N} \left | \arccos (\hat{\bm{w}}_i \cdot \hat{\bm{w}}_j) - \frac{\pi}{2}  \right |$. For a clearer representation, we adopt an angle instead of a radian. It can be seen from Table~\ref{tab:EBV_dim} that dimension has an impact on the performance of a model when training with limited cycles. However, this gap decreases a lot when adopting longer training cycles. Taking $d=100$ and $d=1000$ as an example, the top-1 accuracy under $d=1000$ gains 2.04\%/2.16\% improvement over $d=100$ when training with 105 epochs while this gap is reduced to 0.77\%/0.71\% when training with 305 epochs.

%%%%%%%%%%%%%%%---------------COCO------------%%%%%%%%%%%%%%
\begin{table}[t]
  \centering
  \renewcommand{\arraystretch}{1.0}
  \setlength\tabcolsep{0.1mm}
  \caption{Object detection and instance segmentation on the COCO 2017 dataset. Models are based on Mask R-CNN~\cite{he2017mask} and ``1$\times$" denotes we train models for 12 epochs while ``3$\times$" denotes 36 epochs. ``AP$^{\rm b}$" and ``AP$^{\rm m}$" refer to bounding box AP and mask AP, respectively. Results shaded in gray denote models ending with the general fully connected classifier while others denote models ending with our proposed EBVs.}
    \begin{tabular}{c|c|cccccc}
    \toprule
    Backbone & Schedule & AP$^{\rm b}$   & AP$^{\rm b}_{50}$ & AP$^{\rm b}_{75}$ & AP$^{\rm m}$   & AP$^{\rm m}_{50}$ & AP$^{\rm m}_{75}$ \\
    \hline
    \rowcolor[rgb]{ .851,  .851,  .851} ResNet-50 & 1$\times$ & 38.2  & 58.8  & 41.4  & 34.7  & 55.7  & 37.2   \bigstrut[t]\\
    ResNet-50 (EBVs) & 1$\times$ & \textbf{38.3} & \textbf{59.0} & \textbf{42.0} & \textbf{35.2} & \textbf{56.2} & \textbf{37.7} \\
    \hline
    \rowcolor[rgb]{ .851,  .851,  .851} ResNet-50 & 3$\times$ & 40.9  & 61.3  & 44.8  & 37.1  & 58.3  & 39.9   \bigstrut[t]\\
    ResNet-50 (EBVs) & 3$\times$ & \textbf{41.1} & \textbf{61.7} & \textbf{44.9} & \textbf{37.7} & \textbf{58.9} & \textbf{40.5}  \\
    \hline
    \rowcolor[rgb]{ .851,  .851,  .851} Swin-T & 1$\times$ & 42.7  & 65.2  & 46.8  & 39.3  & \textbf{62.2} & 42.2   \bigstrut[t]\\
    Swin-T (EBVs)  & 1$\times$ & \textbf{42.8} & \textbf{65.3} & \textbf{47.2} & \textbf{39.4} & \textbf{62.2} & \textbf{42.6} \\
    \bottomrule
    \end{tabular}%
  \label{tab:coco}%
\end{table}%

%%%%%%%%%%%%%%-----------ADE20K------------%%%%%%%%%%%%
\begin{table}[t]
  \centering
  \renewcommand{\arraystretch}{1.0}
  \setlength\tabcolsep{1.6mm}
  \caption{Results of semantic segmentation on the ADE20K validation set. Results shaded in gray denote models ending with the general fully connected classifier while others denote models ending with our proposed EBVs. ``1$\times$" denotes we train models for 80,000 steps while ``2$\times$" denotes 160,000 steps.}
    \begin{tabular}{c|c|cc}
    \toprule
    Backbone & Schedule & mIoU  & mIoU(ms+flip)  \\
    \hline
    \rowcolor[rgb]{ .851,  .851,  .851} ResNet-18 & 1$\times$ & 38.76 & 39.81  \bigstrut[t]\\
    ResNet-18 (EBVs) & 1$\times$ & \textbf{38.98} & \textbf{40.09}  \\
    \hline
    \rowcolor[rgb]{ .851,  .851,  .851} ResNet-18 & 2$\times$ & 39.23 & 39.97  \bigstrut[t]\\
    ResNet-18 (EBVs) & 2$\times$ & \textbf{39.75} & \textbf{40.91}  \\
    \hline
    \rowcolor[rgb]{ .851,  .851,  .851} ResNet-50 & 1$\times$ & 40.70  & 41.81  \bigstrut[t]\\
    ResNet-50 (EBVs) & 1$\times$ & \textbf{42.73} & \textbf{44.19}  \\
    \hline
    \rowcolor[rgb]{ .851,  .851,  .851} ResNet-50 & 2$\times$ & 42.05 & 42.78  \bigstrut[t]\\
    ResNet-50 (EBVs) & 2$\times$ & \textbf{42.44} & \textbf{43.94}  \\
    \hline
    \rowcolor[rgb]{ .851,  .851,  .851} Swin-T & 2$\times$ & \textbf{44.41} & 45.79  \bigstrut[t]\\
    Swin-T (EBVs) & 2$\times$ & 44.30  & \textbf{45.88}  \\
    \bottomrule
    \end{tabular}%
  \label{tab:ade20k}%
\end{table}%

\begin{figure*}[t]
\centering
{\includegraphics[width=0.98\textwidth]{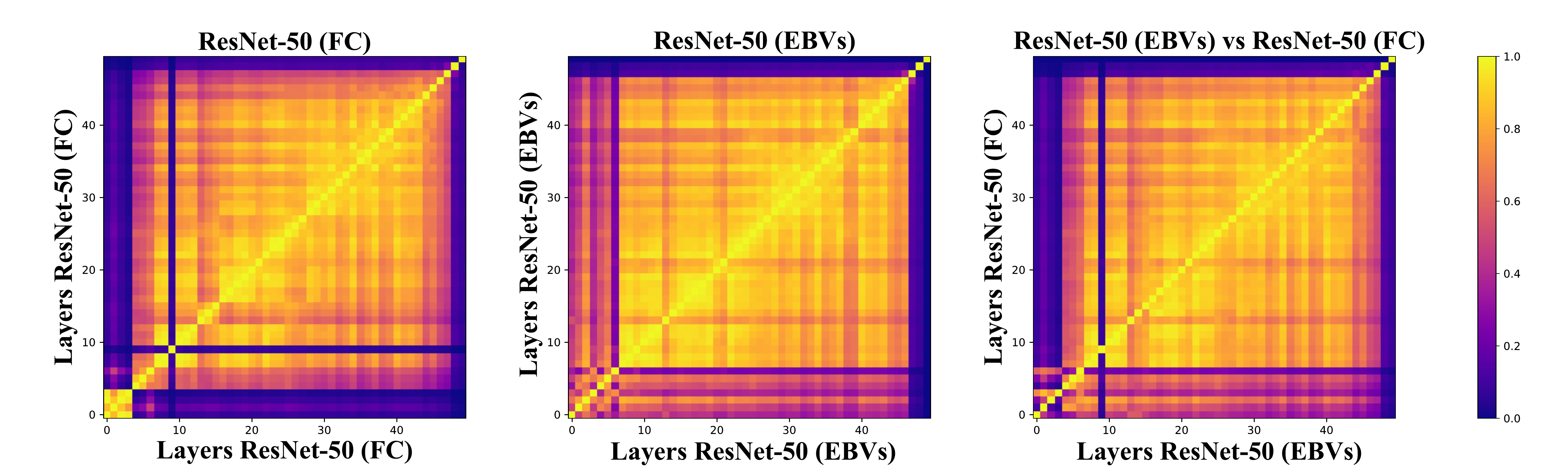}}
\vspace{-0.5em}
\caption{Representation structure of ResNet-50. \textbf{Left:} Similarity between layers within ResNet-50 ends with a general fully connected classifier~(FC). It shows that the last few layers share minimal similarity with the shallow layers while a few shallow layers also share minimal similarity with all the other layers. \textbf{Middle:} Similarity between layers within ResNet-50 ends with EBVs. Only the last few layers share minimal similarity with other layers. \textbf{Right:} Similarity between layers across ResNet-50 ends with a general fully connected layer with \texttt{softmax} and our proposed EBVs. Around 10\% initial layers share a little similarity while the last few layers share the least similarity.}
\label{fig: fcvsebv}
\vspace{-0.8em}
\end{figure*}

\subsection{Empirical evaluations on object detection}
\paragraph{Settings} Object detection and instance segmentation experiments are conducted on the COCO 2017 benchmark~\cite{lin2014microsoft}, which contains 118K images in the training set and 5K images in the validation set. We consider Mask R-CNN~\cite{he2017mask} in MMdetection\cite{chen2019mmdetection} as the detection framework and adopt ResNet~\cite{he2016deep} and Swin Transformer~\cite{liu2021swin} as backbones. For fair comparisons, we utilize the same settings according to MMdetection and previous work. We scaled the image to a maximum of $800$ on the short side and $1333$ on the long side. We adopt 50\% scale horizontal flip and use SGD~\cite{robbins1951stochastic} with momentum$=0.9$ as the optimizer, the initial learning rate is $0.02$. The weight decay equals $0.0001$ while the batch size is $16$ for experiments under the backbone of ResNet. For experiments on Swin Transformer, we use Adam~\cite{kingma2014adam} as the optimizer and the initial learning rate is $0.0001$. The weight decay is changed to $0.05$. To demonstrate the effectiveness of the proposed EBVs, we replace both instance-level and pixel-level classifiers in the detection framework with EBVs. In order to control the parameters of each backbone for fair comparisons, the dimension $d$ of the basis vectors is kept consistent with the number of categories. All backbone models are pre-trained on the ImageNet-1K training set.

\paragraph{Results} Table~\ref{tab:coco} presents comparisons between the general framework and our proposed EBVs. When adopting ResNet-50 as the backbone, the framework ending with EBVs outperforms those ending with fully connected layers among six evaluation indicators. While adopting Swin-T as the backbone, the framework ending with EBVs also achieves comparative results with state-of-the-art performance.

\subsection{Empirical evaluations on semantic segmentation}
\paragraph{Settings} Semantic segmentation experiments are conducted on the ADE20K~\cite{zhou2019semantic}, which contains 20,000 images in the training set and 2,000 images in the validation set. We adopt FPN~\cite{kirillov2019panoptic} and UperNet~\cite{xiao2018unified} in MMSEG~\cite{contributors2020mmsegmentation} as segmentation framework and adopt ResNet~\cite{he2016deep} and Swin Transformer~\cite{liu2021swin} as backbones. For fair comparisons, we utilize the same settings in MMSEG. For experiments on ResNet, we use SGD~\cite{robbins1951stochastic} as the optimizer. The initial learning rate equals $0.01$, weight decay equals $0.0005$, momentum equals $0.9$ and batch size is set as $16$. For experiments on Swin Transformer, we use AdamW~\cite{loshchilov2017decoupled} as the optimizer and the initial learning rate is $0.00006$ while the weight decay is set as $0.01$. To perform our proposed method, We modify pixel-level classifiers in both the decode head and auxiliary head to EBVs. All backbone models are pre-trained on the ImageNet-1K training set.

\paragraph{Results} According to Table~\ref{tab:ade20k}, we find that EBVs surpass the framework ending with fully connected layers as the classifier under the backbone of ResNet. Additionally, when the training step is set as 80,000, EBVs gain a higher mIoU score than the general framework trained for 160,000 steps. While adopting Swin-T as the backbone, the framework ending with EBVs still achieving comparative performance.

\section{Discussions}
We explore whether there are differences in the way EBVs represent and solve image tasks compared to a fully connected layer with \texttt{softmax} in this section. In order to answer this question, we have to analyze the features in the hidden layers as features are usually spread across neurons. However, different layers generally have different numbers of neurons. Recently, Raghu et al.~\cite{raghu2021vision} and Zhen et al.~\cite{zhen2022versatile} applied the Centered Kernel Alignment (CKA)~\cite{cortes2012algorithms,kornblith2019similarity} to solve the above task. CKA is effective because it involves no constraint on the number of neurons. It is also independent of the orthogonal transformations of representations~\cite{zhen2022versatile}. Therefore, we adopt CKA to analyze the above question.

We pick ResNet-50 as the backbone while the experiments on Swin-T can be found in the supplementary materials. One model ends with a general fully connected~(FC) classifier and another model ends with our proposed EBVs. Both of the models are pre-trained on the ImageNet-1K~\cite{deng2009imagenet} training dataset. We use 49 convolutional layers and the last fully connected layer. We plot CKA similarities between all pairs of layers within the whole ImageNet-1K validation dataset in Figure~\ref{fig: fcvsebv}. As shown in Figure~\ref{fig: fcvsebv}, for the ResNet-50 model ending with a general FC classifier, the feature similarity shows that around 10\% initial convolutional layers almost share no similarity with all the other layers while they themselves share a high similarity. If we change the last layer and the learning objective with our proposed EBVs, these shallow layers then share a relatively higher similarity.

\section{Conclusion}
In this paper, we proposed Equiangular Basis Vectors (EBVs) for classification tasks. Different from previous classifiers and classical metric learning methods, EBVs predefined a fixed embedding for all the semantic categories. The learning objective of EBVs then changed to minimize the spherical distance of the embedding of input and each basis vector. Various experiments on the ImageNet-1K~\cite{deng2009imagenet}, COCO 2017~\cite{lin2014microsoft} and ADE20K~\cite{zhou2019semantic} datasets and ablation studies also demonstrated the effectiveness of our proposed EBVs. In the future, we would like to explore relations between each basis vector pair and embed hierarchies when generating the proposed EBVs as the normal Euclidean space could not naturally embed hierarchies on datasets with known semantic hierarchies~\cite{guo2022clipped}. We also would like to explore the performance of EBVs in the case when the number of categories became very large. In addition, EBVs could also be regarded as advanced classifiers, it is interesting to adopt EBVs in other related tasks, \eg, incremental learning, few-shot learning, etc.

\clearpage

%%%%%%%%% REFERENCES
{
\small
\bibliographystyle{ieee_fullname}
\bibliography{EBVbib}
}

\clearpage

\setcounter{section}{0}
\setcounter{table}{0}
\setcounter{figure}{0}
\setcounter{equation}{0}

\onecolumn
\vspace*{24pt}
\begin{center}
    \large
    \lineskip .5em
    \textbf{Equiangular Basis Vectors (Supplementary Materials)}
\end{center}
\vspace*{12pt}

In the supplementary materials, we provide more experiments for the proposed Equiangular Basis Vectors~(EBVs). 

\section{Relations between $\alpha$, $d$ and $N$}
In Section 3.1 of the paper, we make the definition of the proposed EBVs, where $\alpha \in [0,1)$ represents the maximum value of the absolute value of the cosine of the angle between any two vectors, $d$ denotes the dimension of each coordinate vector while $N$ denotes the number of the basis vectors. Specifically, for the EBVs set $\mathcal{W}$, each $\bm{w} \in \mathbb{R}^{d}$ in $\mathcal{W}$ should satisfies:
\begin{equation}\label{eq:_1}
    \forall \bm{w}_i, \bm{w}_j \in \mathcal{W}, i\ne j, \quad 
    -\alpha \leq \frac{\bm{w}_i \cdot \bm{w}_j}{\left \| \bm{w}_i \right \| \left \| \bm{w}_j \right \|} \leq \alpha\,,  
\end{equation}
where $\left \| \cdot \right \| $ denotes the Euclidean norm and ${\rm card}(\mathcal{W})= N$.

According to Elad et al.~\cite{elad2010sparse}, we have known that we can construct a Grassmannian matrix if $N$ satisfies:
\begin{equation}\label{eq:_2}
N < {\rm min}(d(d+1)/2, (N-d)(N-d+1)/2)\,,
\end{equation}
while the lower bound for $\alpha$ equals $\sqrt{\frac{N-d}{d(N-1)}}$. Therefore, we could get a set $\mathcal{W}'$ (${\rm card}(\mathcal{W}')= N$) which satisfies:
\begin{equation}\label{eq:_3}
    \forall \bm{w}_i, \bm{w}_j \in \mathcal{W}', i\ne j, \quad 
    0 \leq \frac{\bm{w}_i \cdot \bm{w}_j}{\left \| \bm{w}_i \right \| \left \| \bm{w}_j \right \|} \leq \alpha\,.  
\end{equation}
However, if $N$ does not satisfy Eq.~\eqref{eq:_2} or the fixed $\alpha$ is larger than the lower bound, we can not construct such a Grassmannian matrix. Furthermore, we would like to explore the relations between $\alpha$, $d$ and $N$. Thus, we use the bisection method to search for the maximum value of $N$ when given fixed $\alpha$ and $d$ which satisfies Eq.~\eqref{eq:_1} according to Algorithm~1 in the paper. In Figure~\ref{fig: adn} in the supplementary materials, we draw the relationship curve between $\alpha$, $d$ and $N$. Specifically, when fixed $\alpha$ and $d$, we calculate a progressive upper bound for $N$. Additionally, it can be easily proved that we can find $n~(2\le n\le N)$ vectors which satisfy Eq.~\eqref{eq:_1} when given the same $\alpha$ and $d$.

\begin{figure}[ht]
\centering
{\includegraphics[width=0.5\textwidth]{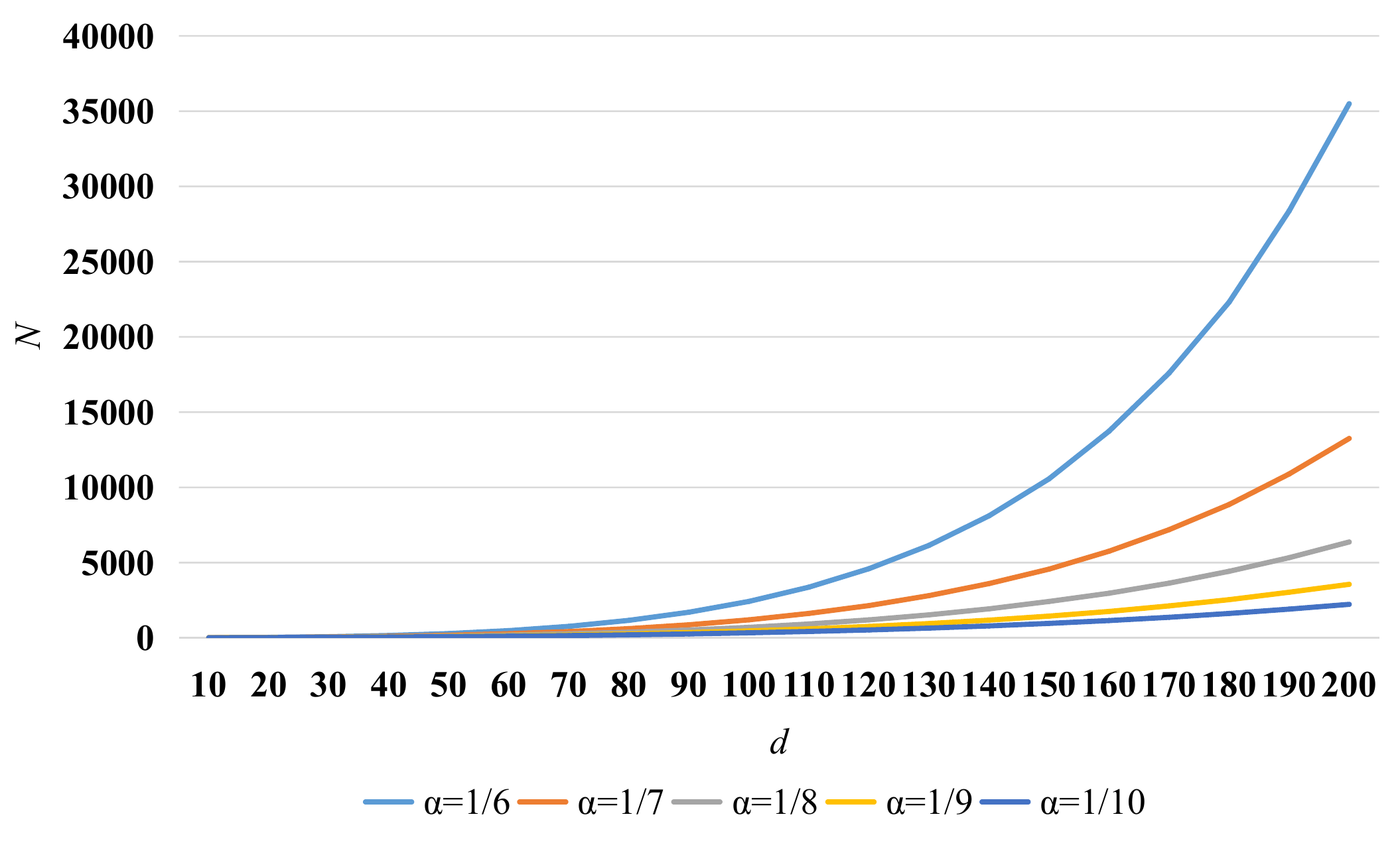}}
\caption{Relations between $\alpha$, $d$ and $N$.}
\label{fig: adn}
\end{figure}

\section{Empirical evaluations on 100,000 classes}
In this section, we conduct experiments in the case where the number of categories reached 100,000.

\begin{table}[htbp]
  \centering
  \small
  \setlength\tabcolsep{1.0mm}
  \caption{Experiments on the dataset with 100,000 classes. ``Params." denotes parameters need to be optimized. ``Top-1 Acc" represents Top-1 accuracy.}
    \begin{tabular}{ccccc}
    \toprule
    Method & Optimizer & EBVs Dim. & Params. (M) & Top-1 Acc ($\%$) \\
    \hline
    FC    & SGD   & --     & 228.4 & 1.29 \bigstrut[t]\\
    FC    & AdamW & --     & 228.4 & \textbf{30.25} \\
    EBVs   & SGD   & 5000  & \textbf{33.8}  & 29.99 \\
    \bottomrule
    \end{tabular}%
  \label{tab:10wclass}%
\end{table}%

\paragraph{Dataset and settings} We collect images containing 100,000 categories with almost the same number of training images as the ImageNet-1K dataset~\cite{deng2009imagenet}. Specifically, we construct a dataset with 100,000 categories, each category contains 12 training images and 6 test images, \ie, a total of 1.2 million images in the training set and 600,000 images in the test set. All these images and labels are collected from the citizen science website iNaturalist\footnote{\url{www.inaturalist.org}}. We adopt ResNet-50 as the backbone and follow Setting A1 in the paper. The hyper-parameters $\tau$ is set as $0.007$ for our EBVs. All the models are pretrained on the ImageNet-1K dataset.

\begin{table*}[htbp]
  \centering
  \caption{Top-1 accuracy of ResNet-32 on the long-tailed CIFAR-10 and CIFAR-100 datasets.}
    \begin{tabular}{c|ccc|ccc}
    \toprule
    Dataset & \multicolumn{3}{c|}{Long-tailed CIFAR-10} & \multicolumn{3}{c}{Long-tailed CIFAR-100} \\
    \hline
    Imbalance ratio & 100   & 50    & 10    & 100   & 50    & 10 \bigstrut[t]\\
    \hline
    FC    & 38.32 & 43.85 & 55.71 & 70.36 & 74.81 & 86.39 \bigstrut[t]\\
    EBVs   & \textbf{40.41} & \textbf{44.68} & \textbf{57.82} & \textbf{73.31} & \textbf{78.97} & \textbf{87.9} \\
    \bottomrule
    \end{tabular}%
  \label{tab:cifar}%
\end{table*}%

\begin{table}[htbp]
  \centering
  \setlength\tabcolsep{5.5mm}
  \caption{Comparisons of classification accuracy ($\%$) on the iNaturalist 2018 dataset.}
  \begin{threeparttable}
    \begin{tabular}{ccc}
    \toprule
    Method & Test size & Top-1 Acc ($\%$) \\
    \midrule
    FC*   & 224$^2$   & 61.7 \\
    FC    & 224$^2$/256$^2$ & 64.03 / 65.43 \\
    EBVs   & 224$^2$/256$^2$ & \textbf{65.00} / \textbf{67.12} \\
    \bottomrule
    \end{tabular}%
    \begin{tablenotes}
        \item[*] denotes the model is trained without TrivialAugment and LR optimizations.
      \end{tablenotes}
    \end{threeparttable}
  \label{tab:inat}%
  \vspace{-1.0em}
\end{table}%

\paragraph{Results} According to Table~\ref{tab:10wclass}, we can see that ResNet-50 ending with a 100,000-way fully connected layer could not work when optimized with SGD~\cite{robbins1951stochastic}. The top-1 accuracy is only 1.29\% after training for 105 epochs. When training with the AdamW~\cite{loshchilov2017decoupled} optimizer, the top-1 accuracy turns out to 30.25\%. However, the 100,000-way fully connected layer contains around 200M parameters which is too large and will become huger if the number of categories continues to grow. When training with our proposed EBVs, if the dimension of each basis vector is set as 5,000, the final top-1 accuarcy gains 29.99\%, while the parameters to be optimized are only 33.8M, which are only around $\frac{1}{7}$ parameters of previous models.

\section{Empirical evaluations on long-tailed image classification}

\subsection{Datasets and settings}

\paragraph{Long-tailed CIFAR-10 \& CIFAR-100} Both CIFAR-10 and CIFAR-100 has 60,000 images of size $32\times 32$ with 50,000 for training and 10,000 for validation. We choose the long-tailed version of CIFAR-10 and CIFAR-100~\cite{cao2019learning}, which downsamples the training data class-wisely from the original dataset by exponential decay functions. For fair comparisons, imbalance factors we use in experiments are $10$, $50$ and $100$.

\paragraph{iNaturalist 2018} iNaturalist 2018~\cite{van2018inaturalist} is a large-scale real-world dataset with 437,513 images from 8,142 categories. It naturally follows a severe long-tailed distribution with an imbalance factor of $512$. Besides the extreme imbalance, it also faces the fine-grained problem~\cite{wei2021fine}. In this paper, the official splits of training and validation images are utilized for fair comparisons. We utilize ResNet-50~\cite{he2016deep} as the backbone.

\paragraph{Settings} 
For long-tailed CIFAR-10 and CIFAR-100 datasets, we follow the data augmentation strategies proposed in~\cite{he2016deep}: randomly crop a $32\times 32$ patch from the original image or its horizontal flip with 4 pixels padded on each side. we use ResNet-32~\cite{he2016deep} as the backbone. SGD optimizer with momentum of $0.9$ and weight decay of $5 \times 10^{-4}$ is used for network optimization. We train all the models for 200 epochs with batch size of $128$. For the iNaturalist 2018 dataset, we utilize ResNet-50~\cite{he2016deep} as the backbone, the hyper-parameters $\tau$ is set as $0.02$. We train the model by following Setting A1 in the paper, the training epoch is set as 200. The dimension of our proposed EBVs is set as 10, 100 and 8,142 for CIFAR-10, CIFAR-100 and iNaturalist 2018, respectively. 

\subsection{Results}
We conduct extensive experiments on long-tailed CIFAR
datasets with three different imbalanced ratios: $10$, $50$ and
$100$. Table~\ref{tab:cifar} reports the top-1 accuracy of models ending with a general $k$-way fully cinnected layer and our proposed EBVs. EBVs outperform the general FC baseline in all the settings. In Table~\ref{tab:inat}, we report the top-1 accuracy on the iNaturalist 2018 dataset. EBVs also gain around 1\% improvement in all the settings.

\begin{table}[b]
  \centering
  \renewcommand{\arraystretch}{1.0}
  \setlength\tabcolsep{1.0mm}
  \caption{Ablation studies of the performance of stacked incremental improvements on top of baseline of our proposed EBVs. w/o EBVs denote models ending with a general fully connected classifier. ResNet-50 baseline is under Setting~A0 in the paper but with only 1-crop testing. ``Top-1 Acc" denotes Top-1 accuracy while ``Abs. Diff." denotes absolute difference. The test size for each image is set as $224^2$ except ``FixRes Mitigations".}
    \begin{tabular}{c|cc}
    \toprule
          & Top-1 Acc (\%) & Abs. Diff. \\
    \hline
    ResNet-50 Baseline & 76.13 & 0.00 \bigstrut[t]\\
    + LR Optimizations w/o EBVs & 76.49 & 0.36 \\
    + TrivialAugment w/o EBVs & 76.81 & 0.68\\
    + TrivialAugment & 77.26 & 1.13 \\
    + Random Erasing & 77.55 & 1.42 \\
    + Label Smoothing & 77.61 & 1.48 \\
    + Mixup & 77.79 & 1.66 \\
    + Cutmix & 78.14 & 2.01 \\
    + Long Training w/o EBVs & 79.51 & 3.38 \\
    + Long Training & 79.73 & 3.60 \\
    + FixRes Mitigations & \textbf{80.90} & \textbf{4.77} \\
    \bottomrule
    \end{tabular}%
  \label{tab:dataaug}%
\end{table}%

\begin{figure*}[t]
\centering
{\includegraphics[width=0.95\textwidth]{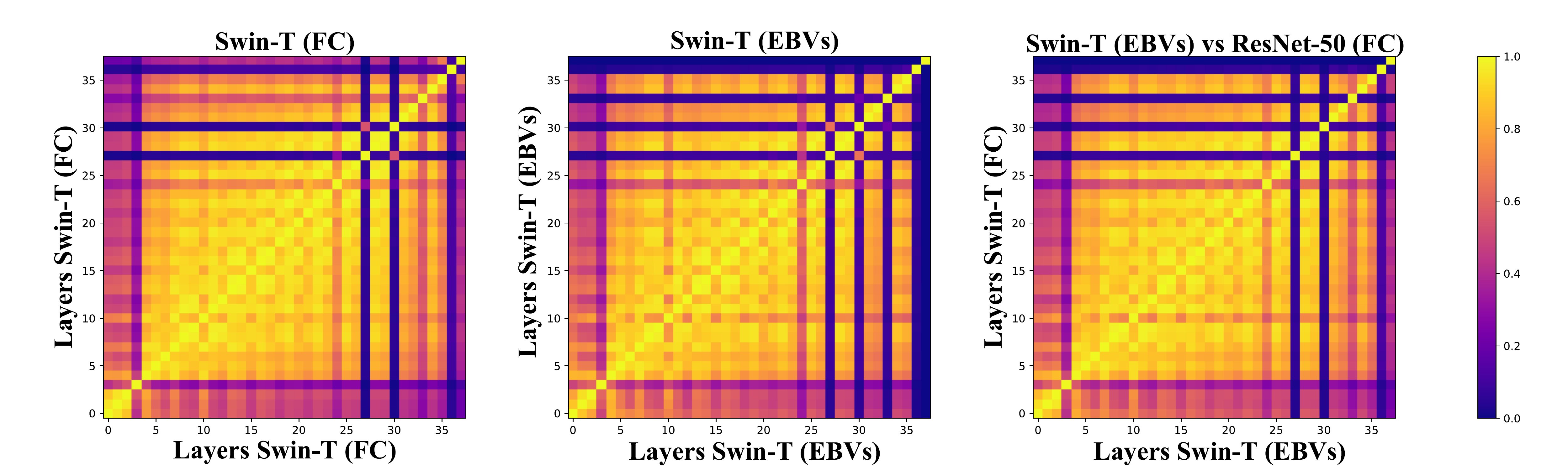}}
\vspace{-1.0em}
\caption{Representation structure of Swin-T. \textbf{Left:} Similarity between layers within Swin-T ending with fully connected layer with softmax. \textbf{Middle:} Similarity between layers within Swin-T ending with EBVs. Only last few layers share minimal similarity with other layers.  \textbf{Right:} Similarity between layers across Swin-T ending with general fully connected layer with softmax and our proposed EBVs. Only last few layers share minimal similarity with other layers.}
\label{fig: fcvsebv2}
\vspace{-1.0em}
\end{figure*}

\section{Ablation studies on training techniques}
In this section, we conduct ablation studies of the performance of different training techniques in our proposed EBVs. As training models are not a journey of monotonically increasing accuracies and the process involves a lot of backtracking~\cite{va2021how}. To quantify the effect of each optimization in our proposed EBVs, we conduct related ablation studies in Table~\ref{tab:dataaug}. When the training crop size is fixed as $224^2$ and turns the inference resolution to $320^2$, with only 1-crop testing, EBVs gains a final top-1 accuracy of 80.9\% on the ImageNet-1K dataset.

\section{Do EBVs perform like FC?}
In this section, we follow Section~5 of the paper and pick Swin-T as the backbone. As shown in Figure~\ref{fig: fcvsebv2} in the supplementary materials, when adopting Swin-T as the backbone, the phenomenon of models ending with EBVs in the last few layers is similar to the performance in ResNet-50. However, most of the other layers share high similarities whether the model ends with a fully connected layer or EBVs.  

\section{Comparisons with methods without a general classifier}
Mettes et al.~\cite{mettes2019hyperspherical} and Yang et al.~\cite{yang2022we} had already employed fixed vectors as learning objectives for the classification task. However, their methods still differs significantly from the approach proposed in this paper. Specifically, Mettes et al.~\cite{mettes2019hyperspherical} proposed using hyperspheres as output spaces, with class prototypes defined a {\emph priori} with large margin separation. Actually, the class prototypes can be considered as the fixed vectors in our proposed EBVs. However, they did not discuss the relationship between the angle, prototype dimension and the number of prototypes. Furthermore, when the prototype dimension is set as $10$, the top-1 acc. on the CIFAR-100 dataset is $20.77\%$ lower than when the prototype dimension is set as $100$, whereas our proposed EBVs only have a $0.97\%$ reduction in top-1 acc. on the ImageNet-1K dataset when the dimension is reduced to $\frac{1}{10}$ of the number of categories. Yang et al.~\cite{yang2022we} studied the potential of learning a neural network for imbalanced classification tasks with the classifier randomly initialized as a simplex equiangular tight frame~(ETF) and fixed during training stage. However, as we have discussed in Section~\ref{sec: pre}, the number of this ETF is linearly correlated with the dimension of equiangular lines as $d \to \infty$. Therefore, their method may not scale well in scenarios with a large number of categories. Overall, our proposed EBVs have advantages over existing methods which have employed fixed vectors in terms of accuracy and scalability.

\end{document}